\documentclass[10pt,twocolumn,letterpaper]{article}
\usepackage[T1]{fontenc}

\usepackage[dvipsnames]{xcolor}

\usepackage{cvpr}
\usepackage{times}
\usepackage{epsfig}
\usepackage{graphicx}
\usepackage{amsmath}
\usepackage{amssymb}

\usepackage[inline]{enumitem}
\usepackage{booktabs}
\usepackage{tabularx}
\usepackage{multirow}
\usepackage{colortbl}
\usepackage{float}
\usepackage{mathtools}

\usepackage{array}
\newcommand{\PreserveBackslash}[1]{\let\temp=\\#1\let\\=\temp}
\newcolumntype{C}[1]{>{\PreserveBackslash\centering}p{#1}}
\newcolumntype{R}[1]{>{\PreserveBackslash\raggedleft}p{#1}}
\newcolumntype{L}[1]{>{\PreserveBackslash\raggedright}p{#1}}

\usepackage{algorithm}
\usepackage{algorithmicx}
\usepackage{algpseudocode}

\usepackage[flushleft]{threeparttable}
\usepackage{comment}
\usepackage{subfig}

\usepackage{algpseudocode,algorithm,algorithmicx}
\usepackage{xspace}
\usepackage{interval}
\usepackage{bm,upgreek}
\usepackage[pagebackref=true,breaklinks=true,letterpaper=true,colorlinks,citecolor=ForestGreen, bookmarks=false]{hyperref}

\usepackage[pagebackref=true,breaklinks=true,letterpaper=true,colorlinks,bookmarks=false]{hyperref}
\graphicspath{{Figures/}}

\newcommand{\clsconf}{\texttt{cls\_c}\xspace}
\newcommand{\locacc}{\texttt{loc\_a}\xspace}

\makeatletter\renewcommand{\paragraph}{%
  \@startsection{paragraph}{4}{\z@}%
                {0.3em \@plus 1ex \@minus 0.2ex}%
                {-0.4em}%
                {\normalfont\normalsize\bf}%
}\makeatother

\cvprfinalcopy %
\def\cvprPaperID{****} %

\ifcvprfinal\fi
\begin{document}
\title{Learning from Noisy Anchors for One-stage Object Detection}

\author{Hengduo Li\thanks{Work partially done while the author was an intern at Salesforce Research.}~~$^{1}$, Zuxuan Wu$^{1}$, Chen Zhu$^{1}$, Caiming Xiong\thanks{Corresponding author.}~~$^{2}$, Richard Socher$^{2}$, Larry S. Davis$^{1}$ \\
$^{1}$University of Maryland \quad $^{2}$Salesforce Research \\
{\tt\small \{hdli,zxwu,chenzhu,lsd\}@cs.umd.edu, \quad \{cxiong,rsocher\}@salesforce.com}
}

\maketitle

\widowpenalty10000
\clubpenalty10000

\begin{abstract}
     State-of-the-art object detectors rely on regressing and classifying an extensive list of possible anchors, which are divided into positive and negative samples based on their intersection-over-union (IoU) with corresponding ground-truth objects. Such a harsh split conditioned on IoU results in binary labels that are potentially noisy and challenging for training. In this paper, we propose to mitigate noise incurred by imperfect label assignment such that the contributions of anchors are dynamically determined by a carefully constructed cleanliness score associated with each anchor. Exploring outputs from both regression and classification branches, the cleanliness scores, estimated without incurring any additional computational overhead, are used not only as soft labels to supervise the training of the classification branch but also sample re-weighting factors for improved localization and classification accuracy. We conduct extensive experiments on COCO, and demonstrate, among other things, the proposed approach steadily improves RetinaNet by $\sim$2\% with various backbones.
\end{abstract}

\thispagestyle{empty}
\section{Introduction}
Object detectors aim to identify rigid bounding boxes that enclose objects of interest in images and have steadily improved over the past few years. Key to the advancement in accuracy is the reduction of object detection to an image classification problem. In particular, a set of candidate boxes, \ie, anchors, of various pre-defined sizes and aspect ratios, are extensively used to be regressed to desired locations and classified into object labels (or background).
While training the regression branch is straightforward with ground-truth (GT) coordinates of objects available, optimizing the classification network is challenging: only a small fraction of anchors sufficiently overlap with GT boxes. This limited number of proposals are considered as positive samples, together with a vast number of remaining negative anchors, to learn good classifiers with the help of techniques like focal loss~\cite{focalloss} or hard sample mining methods~\cite{viola, ohem, libra} that can mitigate data imbalance problems.
\begin{figure}[!t] \centering
  \resizebox{0.93\linewidth}{!}{\includegraphics[width=\linewidth]{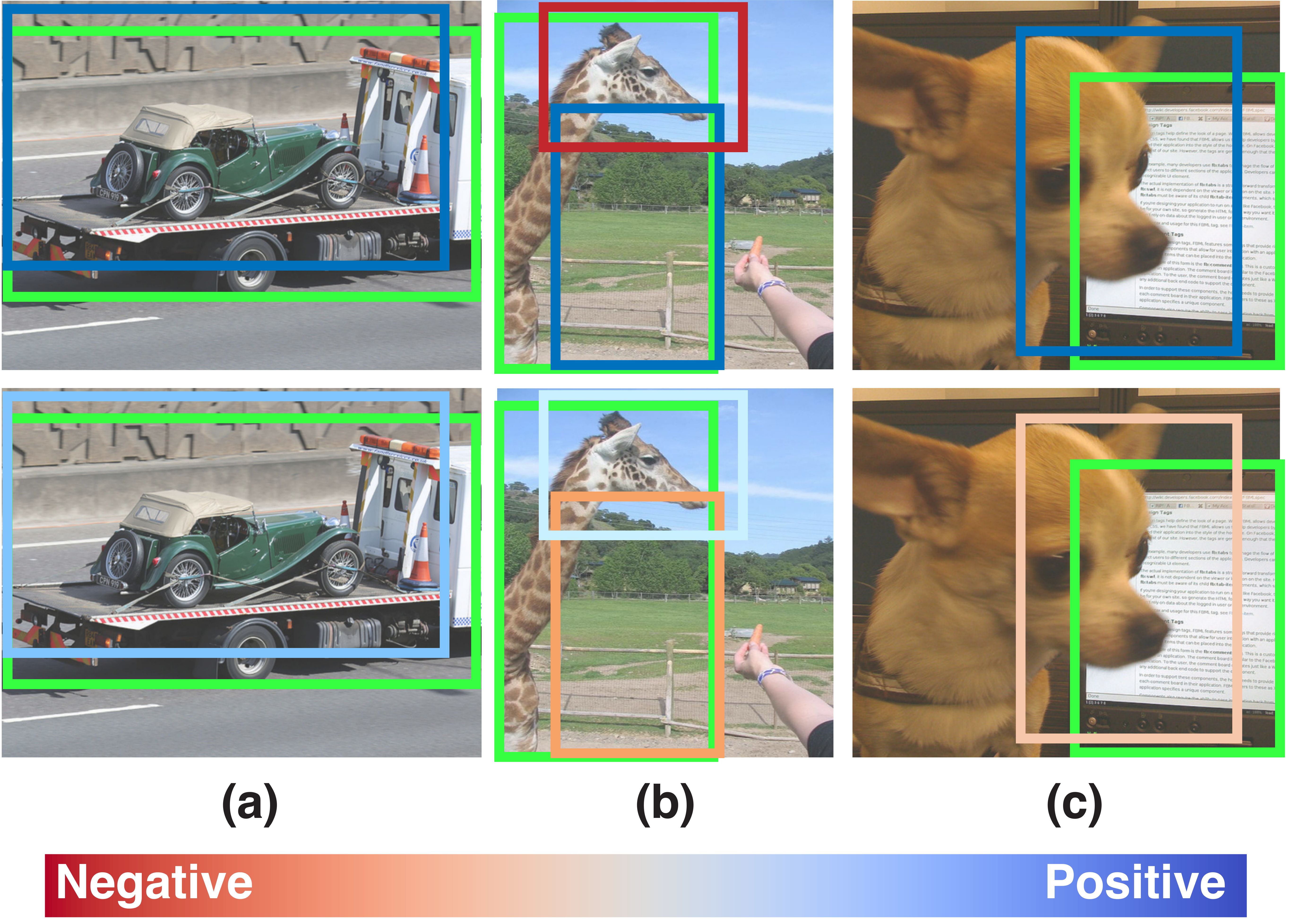}}
  \caption{\textbf{Candidate anchors and their assigned labels (represented by color) in sampled images.} \textbf{Top}: classical training methods assign binary labels, \ie, positive (in blue) and negative (in red) to anchors based on IoU between candidates and ground-truth boxes (in green). \textbf{Bottom}: soft labels, represented by the density of colors, assigned to anchors by our approach based on a proposed cleanliness metric for anchors. Best viewed digitally.}
  \label{fig:teaser}
\end{figure}
 
Despite the success of such training schemes in various detectors~\cite{faster, yolov2, ssd, fpn, focalloss}, the split of positive and negative anchors relies on a design choice---proposals whose IoUs with GT boxes are higher than a pre-defined foreground threshold are considered as positive samples while those with IoUs lower than a background threshold are treated as negative. Although simple and effective, the use of pre-defined thresholds is simply based on ad-hoc heuristics, and more importantly, the resulting hard division of anchors as either positive or negative is questionable. Unlike standard image classification problems where positive and negative samples are more clearly determined by whether an object occurs, anchors that overlap with GT boxes correspond to patches of objects, covering a fraction of an object's extent and thus containing only partial information. Therefore, labels assigned to anchors conditioned on their overlap with GT boxes, are ambiguous. For example, the giraffe head in Figure~\ref{fig:teaser} will be considered as a negative sample since the IoU is low, yet it contains meaningful semantic information useful for both localization and classification. In addition, an axis-aligned candidate with satisfactory overlapping with a GT box might contain background clutter and even other objects (see the green car on the truck and the dog in front of the laptop in Figure~\ref{fig:teaser}), due to the limitations of representing objects using rectangles. Therefore, labels used to train the classification branch are \emph{noisy}, and it is challenging to define perfectly clean labels as there is no oracle information to measure the quality of proposals. In addition, noise in labels is further amplified with sampling methods~\cite{ohem, libra} or focal loss~\cite{focalloss}, since ambiguous and noisy samples tend to produce large losses~\cite{cascade}.  

In light of this, we explicitly consider label noise for anchors with an aim to reduce its impact during classification and regression. In particular, we associate a cleanliness score with each anchor to adaptively adjust its importance during training. Defining cleanliness is non-trivial, since information on the quality of anchors is limited. However, these scores are expected to be (1) determined automatically rather than based on heuristics; (2) soft and continuous so that anchors are not split into positive and negative set with a hard threshold; (3) can reflect the probability of anchors to be successfully regressed to desired locations and classified into object (or background) labels. 

It has been demonstrated that the outputs from networks can indicate the noise level of samples when labels are corrupted and noisy for image classification tasks---the network tends to learn clean samples quickly early on and make confident predictions for them, while recognizing noisy samples slowly yet progressively~\cite{coteaching, mentornet, learntoreweight, combating, learntolearn}. 
In this spirit, we use network outputs as proxies to estimate cleanliness for anchors. We define the cleanliness score of an anchor as a combination of localization accuracies from the regression subnetwork and prediction scores produced by the classification head. Such a definition not only satisfies the aforementioned desiderata but also correlates the classification branch with its regression counterpart. 
This injects localization information to the classification subnetwork, and thus reduces the discrepancies between training and testing, since proposals are simply ranked based on classification confidence with NMS, unaware of localization accuracy during evaluation.

The cleanliness scores then serve as \emph{{soft labels}} to supervise the training of the classification branch. Since they reflect the uncertainty of network predictions and contain richer information than binary labels, this prevents the network from generating over-confident predictions for noisy samples. Furthermore, the cleanliness scores, through a non-linear transformation, are used as \emph{{sample re-weighting}} factors to regulate the contributions of different anchors to loss functions for both classification and regression networks. This assists the model to attend to samples with high cleanliness scores, indicating both accurate regression and classification potentials, and to ignore noisy anchors. It worth pointing out the scores based on the outputs of networks are derived without incurring additional computational cost, and can be readily plugged into anchor-based object detectors. 

We conduct extensive studies on COCO with state-of-the-art one-stage detectors, and demonstrate that our method improves baselines by $\sim$ 2$\%$ using various backbone networks with minimal surgery to loss functions. In particular, with the common practice~\cite{detectron} of multi-scale training, our approach improves RetinaNet~\cite{focalloss} to 41.8\% and a 44.1\% AP with ResNet-101~\cite{resnet} and ResNeXt-101-32$\times$8d~\cite{resnext} as backbones, respectively, which are 2.7\% and 3.3\% higher than the original RetinaNet~\cite{focalloss} and better or comparable with state-of-the-art one-stage object detectors. We additionally show the proposed approach can also be applied to two-stage detectors for improved performance.

\section{Related Work}

\paragraph{Anchor-based object detectors.} Inheriting from the traditional sliding-window paradigms, most modern object detectors perform \textit{classification} and \textit{box regression} conditioned on a set of bounding box priors~\cite{yolov2, ssd, faster, focalloss, cascade, sniper, trident}. In particular, one-stage detectors like RetinaNet~\cite{focalloss}, SSD~\cite{ssd} and YOLOv2~\cite{yolov2} use pre-defined anchors directly, while two-stage detectors like Faster R-CNN~\cite{faster} use generated region proposals refined from anchors either once or in a cascaded manner. A multitude of detectors have been newly proposed based on these frameworks~\cite{rfcn3k, guidedanchoring, gridrcnn, cascade, refinedet, dcn, rfcn, fsaf, htc}.
However, they rely on pre-defined IoU thresholds to assign binary positive and negative labels to proposals in order to train the classification branch. Instead, we associate each box with a carefully designed cleanliness score as soft labels, dynamically adjusting the contributions of different proposals and hence makes the training noise-tolerant. 

\paragraph{Anchor-free object detectors.} There are a few recent studies attempting to address the issues  caused by the use of anchors by formulating object detection as a \textit{keypoint localization} problem. In particular, they aim to localize object keypoints, such as corners~\cite{cornernet}, centers~\cite{centernetzhou, centernetduan} and representative points covering~\cite{fcos} or circumscribing~\cite{reppoints} the spatial extent of objects. The discovered keypoints are either grouped into boxes directly~\cite{cornernet, centernetduan} or used as reference points for box regression~\cite{fcos, reppoints, centernetzhou}. They achieve comparable accuracies with anchor-based counterparts, confirming that the conventional classification supervision using anchors is not perfect.
However, these keypoint-based methods often require more training time to converge. Instead, we improve anchors with slight modifications to loss functions based on the introduced cleanliness scores. This facilitates efficient training yet competitive performance without additional computational cost.

 \paragraph{Sampling/re-weighting in object detection.} The training of object detectors often faces a huge class imbalance due to the large percentage of background candidates. A common technique to address the imbalance is sampling batches with a fixed foreground-to-background ratio~\cite{rcnn, faster}. In addition, various hard~\cite{ohem, libra, dcr, ssd} and soft~\cite{focalloss, ghm, pisa} sampling strategies have been proposed. The core idea of them is to prevent easy samples from overwhelming the loss and then focus the training on hard samples. Despite their effectiveness, these sampling strategies tend to amplify the noise caused by the imperfect split of positive and negative samples, since confusing samples are observed to produce larger losses\cite{coteaching, closerlook}. We demonstrate that our method is complementary to these sampling methods while alleviates the impact of noise for training.

\paragraph{Learning with noisy labels.} 
Extensive studies have been conducted on learning from noisy labels, where noise is generally modeled by deep neural networks~\cite{mentornet,coteaching,combating,learntolearn,learntoreweight} or graphical models~\cite{graph1, graph2}, \etc. Then, outputs from these models are used to re-weight training samples or infer the correct labels. These approaches focus on the task of image classification where noise is from incorrect annotation or caused by the use of weakly-labeled web images from social media or search engines. In contrast, our focus is on object detection, where label noise results from the imperfect split of positive and negative candidates produced by the solely IoU-based label assignment strategy.

\section{Background}
\label{sec:celoss}
We briefly review the standard protocols and design choices for training one-stage detectors and discuss their limitations.
State-of-the-art one-stage detectors take as inputs raw images and produce a set of candidate proposals (\ie, anchors), in the form of feature vectors, to predict the labels of potential objects with a \emph{classification branch}, and regress coordinates of ground-truth bounding boxes through a \emph{regression branch}. In particular, the regression branch typically uses a {smoothed $\ell_1$ loss}~\cite{fast} to encourage correct regression of bounding boxes while the classification counterpart incentivizes accurate predictions of object (or background) labels through a binary cross entropy (\texttt{BCE}) loss~\footnote{We consider binary classification for simplicity, and extending it to multiple classes is straightforward.}: 
\begin{equation} \label{eq:bce}
        \texttt{BCE}\,(p, t) = -t \cdot w_{p} \cdot \texttt{log}(p) - (1-t) \cdot w_{n} \cdot \texttt{log}(1-p),
\end{equation}
where $t \in$ \{0, 1\} denotes the label of a candidate box with background (\texttt{bg}) as 0 and foreground (\texttt{fg}) as 1, and $p \in$ [0, 1] is the predicted classification confidence. $w_{p}$ and $w_{n}$ denote the weighting parameter used in focal loss~\cite{focalloss}, to down-weight well-classified samples. In contrast to standard image classification tasks where labels are more clearly defined based on the presence of objects, the labels of anchors serving as supervisory signals are artificially defined based on their overlapping with GT-boxes in the following way:
\begin{equation} \label{eq:naivestrat}
t = \begin{cases}
        1 &  \text{if } \, \text{IoU} \geq \texttt{fg-threshold} \\
        0 & \text{if } \, \text{IoU} < \texttt{bg-threshold} \\
        -1 & \text{otherwise}.\\
\end{cases}
\end{equation}
The \texttt{fg-threshold} is typically set to $0.5$, which is in part motivated by the PASCAL VOC~\cite{voc} detection benchmark~\cite{ohem} and has been empirically found to be effective for a variety of detectors. Similarly, a box is labeled as background if its IoU with GT is less than the \texttt{bg-threshold}, which is set to $0.4$ in RetinaNet~\cite{focalloss}. 

While offering top-notch performance in most popular detectors, the heuristic approach of identifying positive and negative samples might not be ideal as the thresholds are manually selected and fixed for all objects, regardless of their categories, shapes, sizes, \etc. For example, a candidate box with a high IoU for irregular-shaped objects might contain background clutter or even other objects. On the other hand, anchors with smaller IoUs might still contain important clues. For instance, the candidate box containing a giraffe head in Figure~\ref{fig:teaser} would be considered as background, but it contains useful appearance information for recognizing and localizing a giraffe. This hard label assignment leads to noisy samples which are difficult to learn and produce relatively large losses. As a result, noise will be magnified when re-sampling methods like OHEM~\cite{ohem} or focal loss~\cite{focalloss} are used to mitigate class imbalance and easy sample dominance problems, since more attention is paid to these hard but probably not meaningful proposals.

\section{Our Approach}
As discussed above, noise incurred by the imperfect split of positive and negative samples and the limitations of representing objects with rectangles, not only confuses the classification branch to derive good decision boundaries but also misleads re-sampling/weighting methods. Therefore, we propose to reduce
the impact of noisy proposals by dynamically adjusting their importance. To accomplish this, we introduce the notion of cleanliness for anchors based on their the likelihood to be successfully classified and regressed. Cleanliness scores are continuous in order to adaptively control the contribution of different proposals.
  
Recent advances on learning from noisy labels when training networks suggest that the confidence scores of networks indicates the noise level of samples when making predictions, \ie, networks can easily learn easy samples with high confidence while tending to make uncertain predictions for hard and noisy samples. Motivated by this observation, we define the cleanliness scores of anchors using knowledge learned from the classification and localization branches in detectors:
\begin{equation} \label{eq:sl}
        c = \begin{cases}
                \alpha \cdot  \locacc + (1-\alpha) \cdot \clsconf \quad & \text{for}~b \in \mathcal{A}_{pos} \\
                0  & \text{for}~b \in \mathcal{A}_{neg}. \\
\end{cases}
\end{equation}
Here, $b$ is a candidate box,  \locacc and \clsconf denote the localization accuracy and the classification confidence, respectively, and $\alpha$ is a control parameter, balancing the impact of localization and classification. {In addition, $\mathcal{A}_{pos}$ and $\mathcal{A}_{neg}$ separately represent positive and negative candidate sets from top-\emph{N} proposals for each \texttt{GT}-object based on their IoU before box refinement.} Note that most candidate boxes only cover background regions due to the dense placement of anchors and should not be labeled and learned as positive samples; consequently, we only assign cleanliness scores to a set of plausible positive candidates, with others labeled as 0. Furthermore, we use direct outputs from the classification network as \clsconf and instantiate \locacc as IoU between \emph{regressed} candidate box and its matched \texttt{GT}-object. Note that although we use network outputs, the approach does not suffer from cold start---initial values of \clsconf and output from \emph{regression} branch are both small, so the derived cleanliness score is an approximation of IoU between anchor and matched \texttt{GT}-object, which does not destablize training during the first few iterations.

\paragraph{Soft labels.} The cleanliness scores are readily used as \emph{soft labels} to control the contributions of different anchors to the \texttt{BCE} loss in Equation~\ref{eq:bce} by replacing $t$  with $c$. Since cleanliness scores are dynamically estimated based on the trade-off between \locacc and \clsconf, the network can focus on clean samples and not on improperly labeled noisy samples. In addition, these soft and continuous labels allows the network to be more compatible with detection evaluation protocols, where all final predictions are ranked based on their classification scores in NMS, as will be shown in the experiments. The reasons are two-folds: (1) soft labels prevent the model from generating over-confident binarized decisions, producing more meaningful rankings; (2) the localization accuracies are modeled in the soft labels, reducing the misalignment between classification and localization.

\begin{algorithm}[t]
\footnotesize
\caption{The algorithm of our approach.}
  \textbf{Input:} $\mathcal{I}$, $\mathcal{GT}$, $\mathcal{B}$, \clsconf, \locacc, $\alpha$, $\gamma$, $N$ \\
  $\mathcal{I}$ is the input image, \\
  $\mathcal{GT}$ is the set of ground truth objects within $\mathcal{I}$, \\
  $\mathcal{B}$ is the set of candidates boxes (i.e. anchors), \\
  \clsconf is the classification confidence of corresponding ground truth class for candidates, \\
  \locacc is the localization accuracy of candidates, \\
  $\alpha, \gamma$ are the modulating factors, $N$ controls size of positive candidate set. \\
  \textbf{Output:}  Losses for classification and box regression $L_{cls}, L_{reg}$. 

\begin{algorithmic}[1]
  \State $\mathcal{A}_{pos}, \mathcal{A}_{neg}, S \gets \varnothing$
  \For {$gt \in \mathcal{GT}$}
    \State $indices=argsort($IoU$(\mathcal{B}, gt))$ \Comment{Sort in descending order.}
    \State  $\mathcal{A}_{pos} \gets \mathcal{A}_{pos} \cup \{indices[0:N]: gt\}$
  \EndFor
  \State $\mathcal{A}_{neg} \gets \{(\mathcal{B} - \mathcal{A}_{pos}).indices: 0\}$   

\vspace{1mm}

  \For{$b_i \in \mathcal{A}_{pos}$}
    \State $c = \alpha \cdot \locacc_i + (1-\alpha) \cdot \clsconf_i$ \Comment{Equation~\ref{eq:sl}}
    \State $r = (\alpha \cdot f(\locacc_i) + (1-\alpha) \cdot f(\clsconf_i))^\gamma$ \Comment{Equation~\ref{eq:sr}}
    \State $S \gets S \cup \{b_i: \{c, r\}\}$
  \EndFor  
  \State \textbf{for} $b_i \in \mathcal{A}_{neg}$ \textbf{do}
     $S \gets S \cup \{b_i: \{c \gets 0.0, r \gets 1.0\}\}$ \textbf{end for}
     
\vspace{1mm}
     
  \State $L_{cls} = \Sigma_{i}^{S} r_i \cdot$ \texttt{BCE}$(p_i, c_i)$ \Comment{Equation~\ref{eq:losscls}}
  \State $L_{reg} = \Sigma_{i}^{S} r_i \cdot$ $\texttt{smooth}\_\ell_1$ \Comment{Equation~\ref{eq:lossreg}}

\vspace{1mm}

  \State \textbf{return} $L_{cls}, L_{reg}$
  
\end{algorithmic}
\label{alg:denoise}
\end{algorithm}

\paragraph{Sample re-weighting.} One-stage detectors are usually confronted with a severe imbalance of training data with a large amount of negative proposals and only a few positive ones. To mitigate this issue, focal loss~\cite{focalloss} decreases the loss of easy samples and focuses more on hard and noisy samples. However, for proposals with label noise, they will be stressed during training even though they could simply be outliers. Therefore, we also propose to re-weight samples based on cleanliness scores defined in Equation~\ref{eq:sl}. While we could to directly use Eqn.~\ref{eq:sl} for re-weighting, the variations of cleanliness scores among different proposals are not significantly large as \locacc and \clsconf are normalized. To encourage a large variance, we pass  \locacc and \clsconf through a non-linear function $f(x) = \frac{1}{1-x}$. The re-weighting factor $r$ for each box $b \in \mathcal{A}_{pos}$ becomes:

\begin{equation} \label{eq:sr}
        r = (\alpha \cdot f(\locacc) + (1-\alpha) \cdot f(\clsconf))^\gamma,
\end{equation}
where $\gamma$ is used to further enlarge the score variance, which is fixed to 1 in the experiments.
In addition, we also normalize $r$ to have a mean of 1, since the mean of all positive samples are 1 given that they are equally important before re-weighting. Re-weighting proposals in this way not only downplays the role of very hard samples that cannot be modeled by the network but also helps revisiting clean samples that are regarded as well-classified to promote the discriminative power of classification. 
Finally, with the aforementioned \emph{soft labels} and \emph{sampling re-weighting} factors based on cleanliness scores, loss functions used to train classification $L_{cls}$ and regression $L_{reg}$ networks can then be written as: 

\begin{align}
L_{cls} & = \sum_{i}^{\mathcal{A}_{pos}} r_i~ \texttt{BCE}(p_i, c_i) + \sum_{j}^{\mathcal{A}_{neg}}\texttt{BCE}(p_j, c_j), \label{eq:losscls}\\ 
L_{reg}  & = \sum_{i}^{\mathcal{A}_{pos}} r_i~ \texttt{smooth}\_\ell_1. \label{eq:lossreg}
\end{align}

Here, $r$ is used to weight both losses, \texttt{BCE} loss is computed with $c$ as supervisory signal, and widely adopted smooth $\ell_1$ loss is used for regression~\cite{focalloss}. The complete algorithm of our approach is in Alg.~\ref{alg:denoise}.

\section{Experiments}
\subsection{Experimental Setup} \label{sec:4.1}
\paragraph{Datasets.} We evaluate the proposed approach on the COCO benchmark~\cite{coco}. Following standard training and testing protocols~\cite{focalloss, fpn}, we use the \texttt{trainval35k} set (the union of the $80K$ training images and $35K$ validation images) for training and the \texttt{minival} set ($5K$ images), or the \texttt{test-dev2017} set for testing. The performance is measured by COCO Average Precision (AP)~\cite{coco}. For ablation, we report results on \texttt{minival}. For main results, we report AP on the \texttt{test-dev2017} set where annotations are not publicly available. 

\paragraph{Detectors.} We mainly experiment with RetinaNet~\cite{focalloss}, a state-of-the-art one-stage detector, with different backbones including ResNet-50, ResNet-101~\cite{resnet} and ResNeXt-101-32$\times$8d~\cite{resnext}. In addition, we demonstrate the idea can also be extended to two-stage detectors using Faster R-CNNs~\cite{faster}. For ablation studies, we use RetinaNet with a backbone of ResNet-50.

\paragraph{Implementation details}. We use PyTorch for implementation and adopt 4 GPUs for training with a batch size of 8 (2 images per GPU) using SGD and optimize for $180K$ iterations in total (1$\times$ schedule) unless specified otherwise. The initial learning rate is set as $0.01$ for Faster R-CNNs and $0.005$ for RetinaNet, then divided by $10$ at $120K$ and $160K$ iterations. We use a weight decay of $0.0001$ and a momentum of $0.9$. As in~\cite{focalloss, fcos, fsaf, ghm, pod}, input images are resized to have a shorter side of 800 while the longer side is kept less than 1333; we also perform random horizontal image flipping for data augmentation.  When multi-scale training is performed, input images are jittered over scales \{640, 672, 704, 736, 768, 800\} at shorter side. For multi-scale testing, we use scales \{400, 500, 600, 700, 900, 1000, 1100, 1200\} and horizontal flipping as augmentations following Detectron~\cite{detectron}.

\subsection{Main Results}
We report the performance of our approach on the COCO \texttt{test-dev2017} set using RetinaNet and compare with other state-of-the-art methods in Table~\ref{table:main}. In particular, we compare with variants of RetinaNet such as FSAF~\cite{fsaf}, POD~\cite{pod}, GHM~\cite{ghm}, Cas-Retinanet~\cite{cascaderetina}, RefineDet~\cite{refinedet} and several anchor-free methods including FCOS~\cite{fcos}, Cornernet~\cite{cornernet}, ExtremeNet~\cite{extremenet} and CenterNets~\cite{centernetzhou, centernetduan}. For fair comparisons, following the common setup~\cite{focalloss, fsaf, pod, cascaderetina}, we also train our method with a longer schedule (1.5x of the schedule mentioned in Section~\ref{sec:4.1}) and a scale jittering.

We can see from the table that, without introducing any computational overhead, our method improves RetinaNet by 2.7\% and 3.3\% AP with a ResNet-101 and a ResNeXt-101-32$\times$8d as backbone networks, respectively, confirming the effectiveness of our method. It worth noting all these RetinaNet models are trained with focal loss~\cite{focalloss}, which demonstrates the compatibility of our approach with techniques used to address the imbalance of training samples. In addition, our approach achieves better or comparable performance compared with various state-of-the-art detectors in both single-scale and multi-scale testing scenarios. Note that our approach performs better or on par with some detectors with multiple refinement stages~\cite{refinedet, cascaderetina} or longer training schedule (\eg, a 2x of default schedule)~\cite{fcos, extremenet, cornernet, centernetzhou}. With a strong backbone network ResNeXt-101-32$\times$8d and multi-scale testing, we achieve a high AP of 45.5\%. 

\subsection{Ablation Study}
\paragraph{Different backbone architectures.} We also experiment with different backbone networks for RetinaNet, including ResNet-50, ResNet-101 and ResNeXt-101-32$\times$8d. The results are summarized in Table~\ref{tbl:backbones}. We observe that our method steadily improves the baselines by $\sim$2\% for different backbones.
\begin{table}[!h]
   \centering
   \resizebox{0.95\linewidth}{!}
   {\begin{tabular}{*{7}c}
      \toprule
       Method && Backbone && AP & AP$_{50}$ & AP$_{75}$ \\
      \cmidrule{1-1} \cmidrule{3-3} \cmidrule{5-7}
      Baseline&& \multirow{2}{*}{ResNet-50}  && 36.2 & 54.0 & 38.7  \\
      Ours &&   && \textbf{38.0}{\footnotesize +1.8} & \textbf{56.9} & \textbf{40.6} \\
      \cmidrule{1-1} \cmidrule{3-3} \cmidrule{5-7}

      Baseline &&  \multirow{2}{*}{ResNet-101}  && 38.1 & 56.4 & 40.7 \\
      Ours &&   && \textbf{40.2}{\footnotesize +2.1} & \textbf{59.3} & \textbf{42.9}  \\
      \cmidrule{1-1} \cmidrule{3-3} \cmidrule{5-7}
      Baseline && \multirow{2}{*}{ResNeXt-101}  && 40.3 & 59.2 & 43.1 \\
      Ours &&   && \textbf{42.3}{\footnotesize +2.0} & \textbf{61.6} & \textbf{45.4} \\
      \bottomrule
   \end{tabular}}
   \vspace{-0.1in}
   \caption{\textbf{Results with our approach and comparisons with baselines}, using RetinaNet~\cite{focalloss} with different backbone networks.}
   \label{tbl:backbones}
\end{table}

\begin{table*}[t!] 
    \centering
    \resizebox{\linewidth}{!}{\begin{tabular}{*{11}c}
        \toprule
        Method && Backbone && AP & AP$_{50}$ & AP$_{75}$ && AP$_{S}$ & AP$_{M}$ & AP$_{L}$ \\
        \cmidrule{1-1} \cmidrule{3-3} \cmidrule{5-7} \cmidrule{9-11}
        RetinaNet~\cite{focalloss} && ResNet-101 && 39.1 & 59.1 & 42.3 && 21.9 & 42.7 & 50.2  \\
        Regionlets~\cite{regionlet} && ResNet-101 &&  39.3 & 59.8 & n/a && 21.7 & 43.7 & 50.9 \\
        GHM~\cite{ghm} && ResNet-101 &&  39.9 & 60.8 & 42.5 && 20.3 & 43.6 & 54.1 \\
        FCOS$^\dagger$~\cite{fcos} && ResNet-101 && 41.0 & 60.7 & 44.1 && 24.0 & 44.1 & 51.0  \\
        Cas-RetinaNet~\cite{cascaderetina} && ResNet-101 && 41.1 & 60.7 & 45.0 && 23.7 & 44.4 & 52.9 \\ 
        POD~\cite{pod} && ResNet-101 && 41.5 & 62.4 & 44.9 && 24.5 & 44.8 & 52.9 \\ 
        RefineDet~\cite{refinedet} && ResNet-101 && 36.4/41.8 &  57.5/62.9 & 39.5/45.7 && 16.6/25.6 & 39.9/45.1 & 51.4/54.1  \\
        FSAF~\cite{fsaf} && ResNet-101 &&  40.9/42.8 & 61.5/63.1 & 44.0/46.5 && 24.0/27.8 & 44.2/25.5 & 51.3/53.2 \\
        CenterNet (Duan \etal)$^*\dagger$~\cite{centernetduan} && Hourglass-52  && 41.6/43.5 & 59.4/61.3 & 44.2/46.7 && 22.5/25.3 & 43.1/45.3 & 54.1/55.0 \\ 
        \cmidrule{1-1} \cmidrule{3-3} \cmidrule{5-7} \cmidrule{9-11}

        RetinaNet~\cite{focalloss} && ResNXet-101-32$\times$8d && 40.8 & 61.1 & 44.1 && 24.1 & 44.2 & 51.2  \\
        GHM~\cite{ghm} && ResNXet-101-32$\times$8d && 41.6 & 62.8 & 44.22 && 22.3 & 45.1 & 55.3  \\
        FCOS$^\dagger$~\cite{fcos} && ResNXet-101-32$\times$8d && 42.1 & 62.1 & 45.2 && 25.6 & 44.9 & 52.0  \\
        FSAF~\cite{fsaf} && ResNXet-101-32$\times$8d && 42.9/44.6 & 63.8/65.2 & 46.3/48.6 && 26.6/29.7 & 46.2/47.1 & 52.7/54.6  \\
        CornerNet$^*\dagger$~\cite{cornernet} && Hourglass-104 && 40.5/42.1 & 56.5/57.8 & 43.1/45.3 && 19.4/20.8 & 42.7/44.8 & 53.9/56.7\\
        ExtremeNet$^*\dagger$~\cite{extremenet} && Hourglass-104 && 40.2/43.7 & 55.5/60.5 & 43.2/47.0&&  20.4/24.1 & 43.2/46.9 & 53.1/57.6 \\
        CenterNet (Zhou \etal)$^*\dagger$~\cite{centernetzhou} && Hourglass-104 & & 42.1/45.1 & 61.1/63.9 & 45.9/49.3 && 24.1/26.6 & 45.5/47.1 & 52.8/57.7 \\
        CenterNet (Duan \etal)$^*\dagger$~\cite{centernetduan} && Hourglass-104 & & 44.9/47.0 & 62.4/64.5 & 48.1/50.7 && 25.6/28.9 & 47.4/49.9 & 57.4/58.9 \\
        \cmidrule{1-1} \cmidrule{3-3} \cmidrule{5-7} \cmidrule{9-11}

        \textbf{Ours} && ResNet-101  && 41.8/43.4 & 61.1/62.5 & 44.9/47.0 && 23.4/26.0 & 44.9/46.0 & 52.9/55.4  \\
        \textbf{Ours} && ResNXet-101-32$\times$8d && 44.1/45.5 & 63.8/65.0 & 47.5/49.3 && 26.0/28.2 & 47.4/48.4 & 55.0/57.6 \\
        \bottomrule
    \end{tabular}}
    \begin{tablenotes} \small \item $*$ Horizontal flipping used for both single-scale and multi-scale testing
    \item $\dagger$ Longer training schedule
    \end{tablenotes}
    \vspace{-0.1in}
    \caption{Detection results (\% AP) on COCO \texttt{test-dev2017} set. Single-scale / multi-scale (if exists) testing results are reported. Our method improves RetinaNet detectors by $\approx$ $3\%$ AP and obtains better or comparable performance compared with state-of-the-art one-stage detectors.}
    \label{table:main}
    \end{table*}

\paragraph{Contributions of soft labels (SL) and re-weighting (SR).} To demonstrate the effectiveness of the two key components based on cleanliness scores, we report the results of our approach using SL and SR, separately in Table~\ref{table:ab_components}. We can see that applying either \emph{soft labels} or \emph{re-weighting coefficients} derived from the cleanliness scores improves the baselines, while combining both methods offers the largest performance improvement. It worth pointing out when \emph{soft labels} are not applied, simply \emph{re-weighting} the samples with hard binary samples brings relatively minor performance gain, suggesting the use of soft supervisory signals for training the classification branch is critical.
\begin{table}[h!]
\centering
\resizebox{1.0\linewidth}{!}{
 \begin{tabular}{*{11}c}
  \toprule
   \textbf{SL} &&  \textbf{SR} && AP & AP$_{50}$ & AP$_{75}$ && AP$_{S}$ & AP$_{M}$ & AP$_{L}$  \\
 \cmidrule{1-1} \cmidrule{3-3} \cmidrule{5-7} \cmidrule{9-11}
     && && 36.2 & 54.0 & 38.7 && 19.3 & 40.1 & 48.8 \\
    \checkmark &&  && 37.1 & 56.5 & 40.0 && 19.4 & 40.9 & 49.3\\
    && \checkmark && 36.7 & 54.4 & 39.3 && 19.5 & 40.3 & 49.4 \\
    \checkmark && \checkmark && \textbf{37.7} & \textbf{56.5} & \textbf{40.2} && \textbf{20.0} & \textbf{41.1} & \textbf{51.2}\\
 \bottomrule
 \end{tabular}}
 \vspace{-0.1in}
 \caption{Ablation experiments on the effectiveness of components in our method, Soft Labels (\textbf{SL}) and Sample Re-weighting (\textbf{SR}). }
 \label{table:ab_components}
\end{table}

\begin{table*}[!t] \centering
\subfloat[Varying $\gamma$ for sample re-weighting.]{\resizebox{!}{1.6cm}{\begin{tabular}{*{5}c}
      \toprule
      $\gamma$ && AP & AP$_{50}$ & AP$_{75}$  \\
      \cmidrule{1-1} \cmidrule{3-5}
      0.0 && 37.1 & 56.6 & 40.0  \\
      0.5 && 37.6 & 56.9 & 40.1  \\
      1.0 && 37.7 & 56.5 & 40.2  \\
      1.25 && 37.7 & 56.2 & 40.3 \\
      1.5 && 37.7 & 55.9 & 40.5  \\
      1.75 && 35.9 & 52.9 & 38.4  \\
      \bottomrule
  \end{tabular}}} \hspace{6mm}
\subfloat[Varying $N$ for collecting $\mathcal{A}_{pos}$.]{\resizebox{!}{1.6cm}{\begin{tabular}{*{7}c}
   \toprule
      $N$ && $\gamma$ && AP & AP$_{50}$ & AP$_{75}$  \\
      \cmidrule{1-1} \cmidrule{3-3} \cmidrule{5-7}

      30 && 1.0 && 37.7 & 56.5 & 40.2  \\
      40 && 1.0 && 37.7 & 56.8 & 40.3  \\
      50 && 1.0 && 37.3 & 56.2 & 39.9  \\
      60 && 1.0 && 37.1 & 55.8 & 39.5  \\
      80 && 1.0 && 36.6 & 55.6 & 38.9  \\
      80 && 1.25 && 36.9 & 55.5 & 39.2 \\
   \bottomrule
  \end{tabular}}} \hspace{6mm}
\subfloat[Varying the balancing factor $\alpha$.]{
\label{tbl:alpha}
\renewcommand{\arraystretch}{1.2}
\resizebox{!}{1.6cm}{\begin{tabular}{*{5}c}
   \toprule
      $ \alpha$ && AP & AP$_{50}$ & AP$_{75}$  \\
      \cmidrule{1-1} \cmidrule{3-5}
      0.0 && 37.3 & 56.3 & 39.7  \\
      0.25 && 37.3 & 56.2 & 39.9  \\
      0.5 && 37.7 & 56.5 & 40.2  \\
      0.75 && 38.0 & 56.9 & 40.6  \\
      1.0 && 37.8 & 56.5 & 40.5 \\
   \bottomrule
  \end{tabular}}} \\
\caption{\textbf{Ablation experiments on sensitivity of hyper-parameters in our method}: (a) $\gamma$ modulates the degree of focus on different samples. (b) $N$ controls size of $\mathcal{A}_{pos}$. (c) $\alpha$ balances \clsconf and \locacc in calculating cleanliness score.} \label{table:ablation}
\end{table*}

\paragraph{Hyper-parameters sensitivity.} We also analyze the sensitivity of different hyper-parameters used in our approach: $\gamma$ controls the degree of focus on different samples, $N$ governs the size of $\mathcal{A}_{pos}$ and $\alpha$ balances \clsconf and \locacc when computing cleanliness scores. 
As shown in Table~\ref{table:ablation}, our method is relatively robust to different parameters. 
We observe that $\gamma$ and $N$ should be selected together, since a large $\gamma$ focuses training on a small proportion of samples while a large $N$ adds more noisy samples to $\mathcal{A}_{pos}$; detection performance would drastically degrade nevertheless if either of them is too large. When $\gamma=0$, all samples are equally re-weighted for network learning and SR is thus disabled. The effect of $\alpha$ reveals the trade-off between \clsconf and \locacc to compute cleanliness scores for label assignment and sample re-weighting. As shown in Table~\ref{tbl:alpha}, $\alpha=0.75$ yields the best result---\locacc tends to be more important than \clsconf as larger $\alpha$ offers better performance. This also confirms that considering both classification and regression branches when defining the cleanliness scores is important. 

\paragraph{Extension to two-stage detectors.} Our method offers clear performance gains for one-stage detectors, and we hypothesize that it could be easily plugged into multi-stage detectors, producing better proposals. We validate our assumption with Faster R-CNN~\cite{faster}. In particular, we first train the Region Proposal Network (RPN) with our approach to analyze recalls, since one-stage detectors are a variant of RPN. Table~\ref{table:ab_rpn} presents the recall of generated proposals with different methods. We can see that our approach outperforms the baseline RPN model by clear margins---7.8, 5.4, 3.4 percentage points for AR$_{100}$, AR$_{300}$ and AR$_{1000}$, respectively. It also surpasses the performance of a two-stage iterative RPN in~\cite{refinedet} and a ``RefineRPN'' structure similar to~\cite{refinedet} where anchors are regressed and classified twice with different features. Note that larger improvements are observed when a smaller number of proposals are kept, suggesting that our method can be better at ranking predictions according to actual localization accuracy.
We also analyze the contributions of soft labels and sample re-weighting, and observe similar trends as in one-stage detectors.

\begin{table}[!h] \centering
    \resizebox{0.9\linewidth}{!}{
    \begin{tabular}{*{5}c}
     \toprule
       Method && AR$_{100}$ & AR$_{300}$ & AR$_{1000}$   \\
       \cmidrule{1-1} \cmidrule{3-5}
       RPN Baseline~\cite{fpn} && 43.3 & 51.6 & 56.9 \\
       RPN-0.5 && 46.8 & 53.4 & 56.2 \\
       RPN+Iterative~\cite{guidedanchoring} && 49.7 & 56.0 & 60.0 \\
       RefineRPN~\cite{guidedanchoring, refinedet} && 50.2 & 56.3 & \textbf{60.6} \\
      \cmidrule{1-1} \cmidrule{3-5}
       RPN-0.5 + SR && 48.3 & 54.6 & 56.6\\ 
       Ours && \textbf{51.1} & \textbf{57.0} & \textbf{60.3} \\
    \bottomrule
    \end{tabular}}
    \vspace{-0.1in}
    \caption{\textbf{Results of region proposals} evaluated on COCO \texttt{minival}, measured by Average Recall (AR). RPN Baseline uses \{0.3, 0.7\} IoU thresholds for GT assignment (background if $<$0.3, foreground if $>$0.7, ignored if in between) while RPN-0.5 uses to separate positive and negative samples. SR denotes sample reweighting.}
    \label{table:ab_rpn}
   \end{table}
   
\begin{table}[!h] \centering
\resizebox{\linewidth}{!}{
    \begin{tabular}{*{9}c}
    \toprule
    Method && AP & AP$_{50}$ & AP$_{75}$ && AP$_{S}$ & AP$_{M}$ & AP$_{L}$   \\
    \cmidrule{1-1} \cmidrule{3-5} \cmidrule{7-9}
    Baseline && 36.8 & 58.5 & 39.8 && 21.0 & 39.9 & 47.6 \\
    Ours && \textbf{37.8} & \textbf{59.2} & \textbf{41.1} && \textbf{21.7} & \textbf{41.3} & \textbf{48.9}  \\
    \bottomrule
\end{tabular}}
\vspace{-0.1in}
\caption{\textbf{Results of Faster R-CNN with FPN}, with and without our approach.} \label{table:ab_frcnn}
\end{table}

We then train a Faster R-CNN~\cite{faster} with FPN~\cite{fpn} in an end-to-end manner by using our approach only for RPNs. The results are shown in Table~\ref{table:ab_frcnn}. We observe 1\% mAP improvement compared to standard training of faster rcnns, demonstrating that our approach is also applicable to two-stage detectors without any additional computation.

\subsection{Discussions}
 In this section, we perform various quantitative and qualitative analyses to investigate the performance gains brought by our approach.
 
 \begin{figure}[!h] \centering
   \resizebox{1.0\linewidth}{!}{\includegraphics[width=\linewidth]{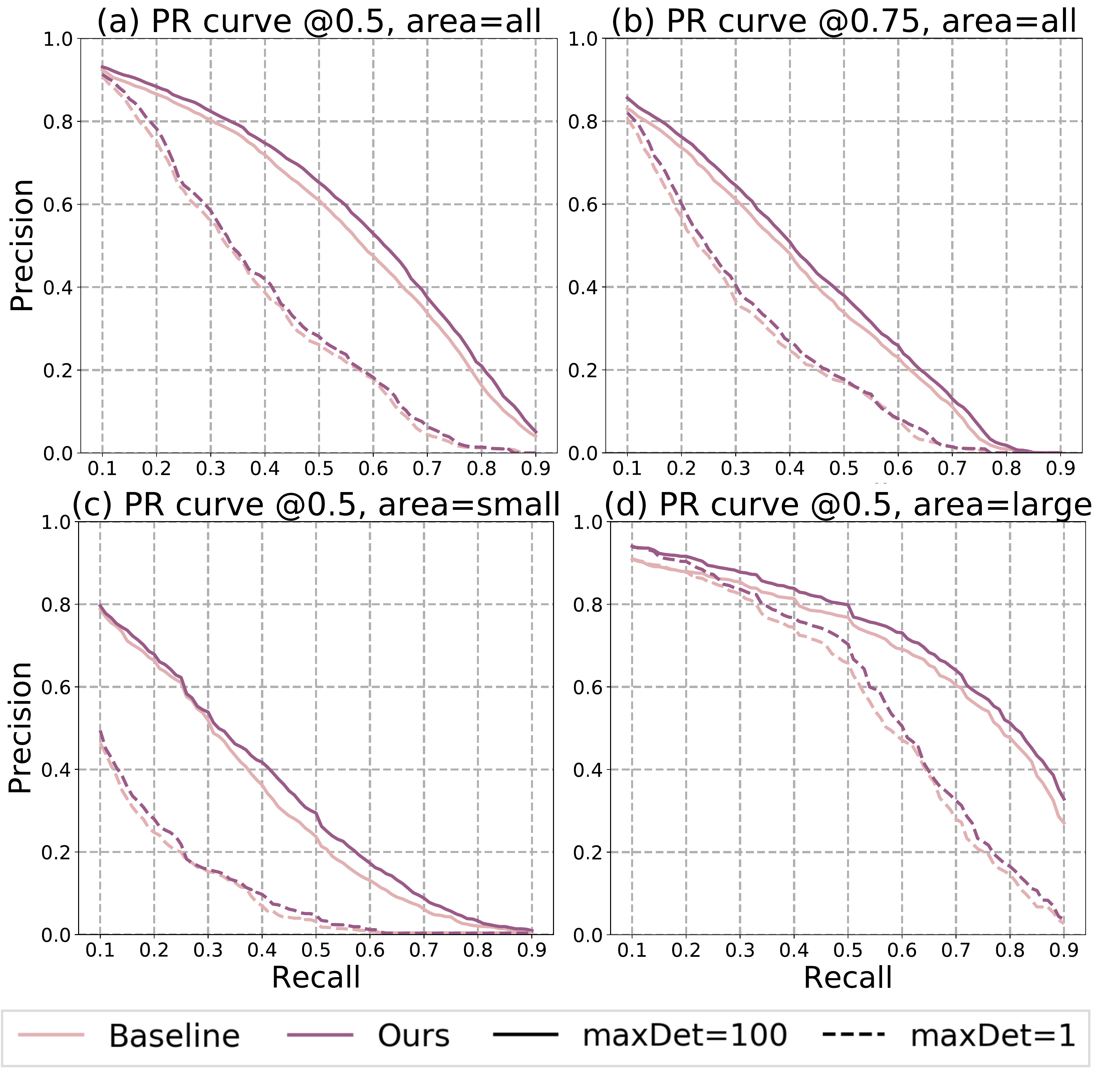}}
   \vspace{-0.3in}
   \caption{\textbf{Precision \vs recall (PR) curves} of our approach, and comparisons with baselines under different IoU thresholds, object sizes (area) and maximum number of predictions per image (maxDet).}
   \label{fig:pr_curve}
 \end{figure}

 \paragraph{Recall \vs precision.} To better understand how our method improves detection performance, we plot the precision \vs recall curves in Fig~\ref{fig:pr_curve} and analyze the performance gains. As demonstrated, our method steadily promotes detection performance in different conditions like IoU thresholds, object sizes and maximum number of predictions per image during evaluation. It is also worth noting that our method obtains clear precision gains for all recall ratios, and hence it could be beneficial to various object detection applications in real-world scenarios.
 
 \begin{figure*}[!t] \centering
   \resizebox{0.9\linewidth}{!}{\includegraphics[width=\linewidth]{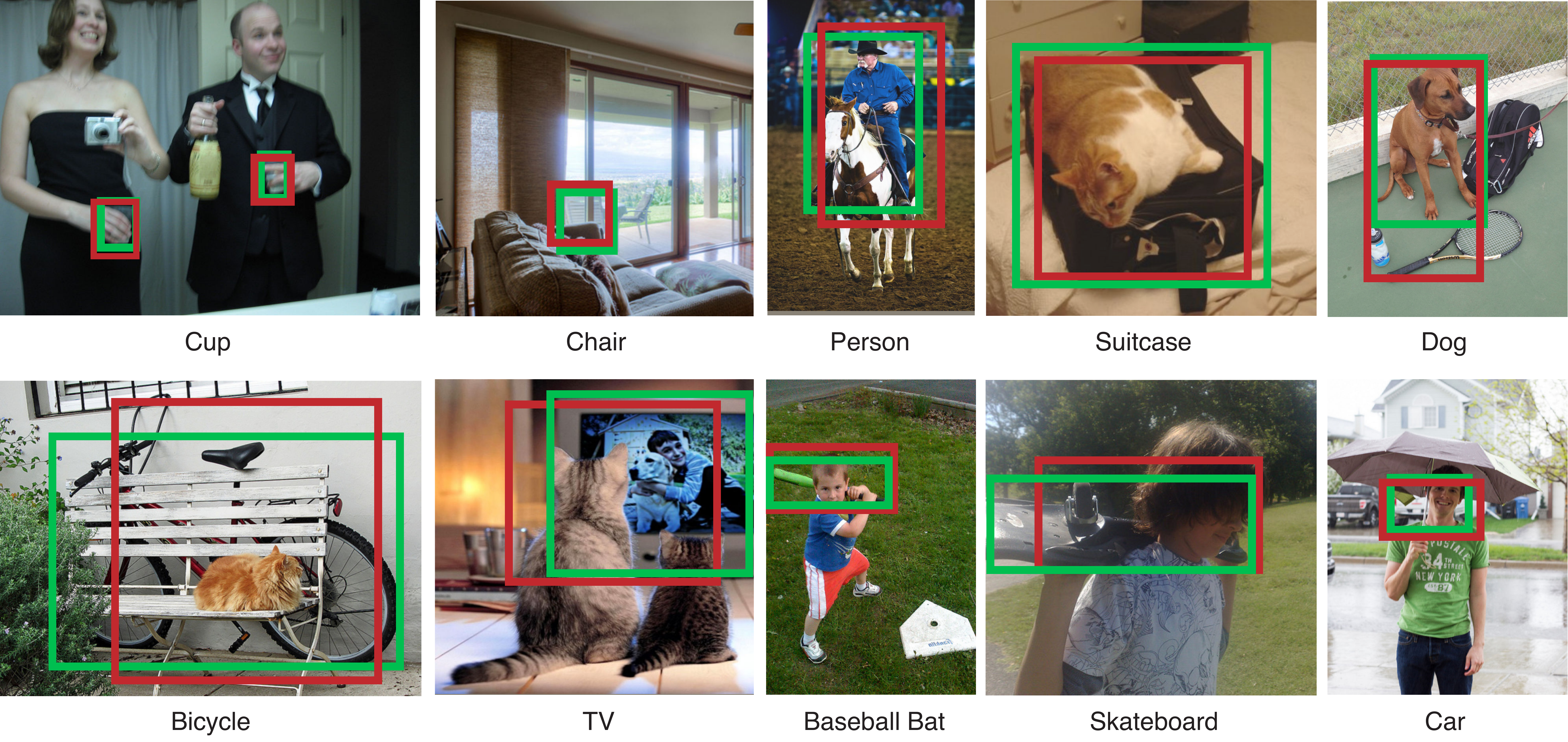}}
   \vspace{-0.1in}
   \caption{\textbf{Example candidate boxes (in red) with high IoUs with ground-truth boxes (in green), yet containing noisy contents, are down-weighted by our method.} Anchors and ground-truth boxes are denoted in red and green, respectively.}
   \label{fig:qual}
 \end{figure*}

 \paragraph{Classification confidence prediction.} 
 We also analyze the predicted classification confidence and investigate whether our proposed method helps alleviate the issue of over confident predictions and reduce the discrepancy between classification prediction and localization accuracy. For baseline detectors and our method, {we collect their top-2\% confident predictions} on COCO \texttt{minival} set before and after NMS and then calculate their mean classification confidence and IoUs with matched ground-truth boxes. As shown in Table~\ref{table:ana}, detectors trained with our method produces relatively milder predictions than the baseline for classification. Although predictions of the baseline offer a higher average IoU before NMS, it is surpassed by our method after running NMS. This suggests that our method is more friendly when ranking is performed during evaluation since predicted labels are softer and contain more ordering information, and thus is more compatible with NMS. To further verify the ability of our method to correlate classification confidence with localization accuracy, we calculate the Pearson correlation coefficient on these predictions before NMS, and the coefficients between classification confidence and output IoU are 0.169 and 0.194 for baseline and our approach, respectively. This indicates that cleanliness scores considering both branches are able to help bridge the gap between classification and localization.
 
 \begin{table}[!h]  \centering
  \resizebox{1.0\linewidth}{!}{\begin{tabular}{*{7}c}
   \toprule
    \multirow{2}{*}{Method}    && \multicolumn{2}{c}{Before NMS} && \multicolumn{2}{c}{After NMS} \\
    \cmidrule{3-4} \cmidrule{6-7}

     &&  Mean Conf & Mean IoU && Mean Conf & Mean IoU \\
   \cmidrule{1-1} \cmidrule{3-4} \cmidrule{6-7}
    Baseline && 0.845 & 0.895 && 0.958 & 0.914\\
    Ours && 0.782 & 0.882 && 0.920 & 0.921\\ 
  \bottomrule
  \end{tabular}}
  \vspace{-0.1in}
  \caption{\textbf{Mean classification confidence and output IoUs} with matched ground-truth using predictions before (\textbf{left}) and after NMS (\textbf{right}).}
  \label{table:ana}
 \end{table}
 
 \paragraph{Qualitative analysis.} In addition to quantitative results, we also demonstrate qualitatively in Figure~\ref{fig:qual} that our method is able to down-weight noisy anchors. As shown in the Figure, our method assigns smaller soft label and re-weighting coefficients to ambiguous samples that contain irrelevant objects or complex background. For example, the anchors encompassing cups in the top-left of Figure~\ref{fig:qual} are occluded by the lady's and gentleman's hands and thus are down-weighted, although they sufficiently overlap with the ground-truth. Similarly, the anchor in the top-middle associated with the person is also down-weighted, since it largely contains irrelevant regions from a horse. This verifies that the label noise can be modeled by our definition of cleanliness and hence are mitigated to improve the training process of object detection.  We note that these ambiguous anchors are fairly common---such anchors can be easily found across ten different categories as shown in Figure~\ref{fig:qual}.

\section{Conclusion}
In this paper, we have presented an approach that is explicitly designed to mitigate noise in anchors used for training object detectors. In particular, we introduced a carefully designed cleanliness score for each anchor used to dynamically adjust their importance during training. These cleanliness scores, leveraging outputs from classification and detection branches, serve as proxies to measure the probability of anchors to be successfully regressed and classified. They are further used as soft supervisory signals to train the classification network and re-weight samples to achieve better localization and classification performance. Extensive studies have been conducted on COCO, and the results demonstrate the effectiveness of the proposed approach both quantitatively and qualitatively.

\footnotesize{\noindent\textbf{Acknowledgement} HL, ZW and LSD are supported by the Intelligence Advanced Research Projects Activity (IARPA) via Department of Interior/Interior Business Center (DOI/IBC) contract number D17PC00345. The U.S. Government is authorized to reproduce and distribute reprints for Governmental purposes not withstanding any copyright annotation thereon. The views and conclusions contained herein are those of the authors and should not be interpreted as necessarily representing the official policies or endorsements, either expressed or implied of IARPA, DOI/IBC or the U.S. Government.}

{\small
\bibliographystyle{ieee_fullname}
\bibliography{reference}
}

\end{document}